\documentclass[nohyperref]{article}

\usepackage{microtype}
\usepackage{graphicx}
\usepackage{subfigure}
\usepackage{booktabs} 

\usepackage{hyperref}

\usepackage[accepted]{icml2023}

\usepackage{amsmath}
\usepackage{amssymb}
\usepackage{mathtools}
\usepackage{amsthm}

\usepackage[capitalize,noabbrev]{cleveref}

\usepackage{makecell}
\usepackage{multirow}
\usepackage{tabularx}

\theoremstyle{plain}

\theoremstyle{definition}

\theoremstyle{remark}

\usepackage[textsize=tiny]{todonotes}

\icmltitlerunning{Clinical Decision Transformer}

\begin{document}

\twocolumn[
\icmltitle{Clinical Decision Transformer: Intended Treatment \\
Recommendation through Goal Prompting}



\icmlsetsymbol{equal}{*}

\begin{icmlauthorlist}
\icmlauthor{Seunghyun Lee}{equal,postech}
\icmlauthor{Da Young Lee}{equal,kucm}
\icmlauthor{Sujeong Im}{postech}
\icmlauthor{Nan Hee Kim}{kucm}
\icmlauthor{Sung-Min Park}{postech}
\end{icmlauthorlist}

\icmlaffiliation{postech}{Department of Convergence IT Engineering, Pohang University of Science and Technology (POSTECH), Pohang, Korea}
\icmlaffiliation{kucm}{Division of Endocrinology and Metabolism, Department of Internal Medicine, Korea University College of Medicine, Seoul, Korea}

\icmlcorrespondingauthor{Sung-Min Park}{sungminpark@postech.ac.kr}
\icmlcorrespondingauthor{Nan Hee Kim}{nhkendo@gmail.com}

\icmlkeywords{Clinical Recommender System, Electronic Health Records, Foundation Model, Causal Inference}

\vskip 0.3in
]



\printAffiliationsAndNotice{\icmlEqualContribution} 

\begin{abstract}
With recent achievements in tasks requiring context awareness, foundation models have been adopted to treat large-scale data from electronic health record (EHR) systems. However, previous clinical recommender systems based on foundation models have a limited purpose of imitating clinicians’ behavior and do not directly consider a problem of missing values. In this paper, we propose Clinical Decision Transformer (CDT), a recommender system that generates a sequence of medications to reach a desired range of clinical states given as goal prompts. For this, we conducted goal-conditioned sequencing, which generated a subsequence of treatment history with prepended future goal state, and trained the CDT to model sequential medications required to reach that goal state. For contextual embedding over intra-admission and inter-admissions, we adopted a GPT-based architecture with an admission-wise attention mask and column embedding. In an experiment, we extracted a diabetes dataset from an EHR system, which contained treatment histories of 4788 patients. We observed that the CDT achieved the intended treatment effect according to goal prompt ranges (e.g., NormalA1c, LowerA1c, and HigherA1c), contrary to the case with  behavior cloning. To the best of our knowledge, this is the first study to explore clinical recommendations from the perspective of goal prompting. See \href{https://clinical-decision-transformer.github.io}{https://clinical-decision-transformer.github.io} for code and additional information. 
\end{abstract}

\section{Introduction}

As electronic health record (EHR) systems have become common in large medical organizations, it is not unusual for the system to maintain many years of diagnoses and treatment histories from millions of patients as large-scale computerized data. With the recent increase in the size of both models and data in machine learning, an increasing number of studies are attempting to utilize EHR data for various purposes, such as clinical insights \cite{acosta2022multimodal}, treatment effect estimation \cite{ling2022emulate}, patient representation \cite{solares2021transfer}, and clinical decision-support \cite{sutton2020overview}.

In utilizing the clinical histories of individuals, contextual embedding poses a significant challenge. EHR data frequently contain missing values in laboratory test results for several reasons, such as prior knowledge about an individual’s condition or relative clinical priority. \citet{che2018recurrent} note this omission as informative missingness based on the analysis that the missing rate can be correlated to mortality and diagnosis categories. Thus, missing data must be carefully considered without exclusion, and in this process, it is necessary to infer the underlying information intimately through the context surrounding the missing points. Moreover, as the individual treatment and response history is highly heterogeneous across patients, recent studies and position statements from medical associations have demonstrated that there is no one-size-fits-all treatment \cite{subramanian2014personalized}. For this reason, understanding heterogeneity in the clinical context is crucial in realizing individual-oriented usage of recorded health data from various patients. Therefore, contextual embedding for informative missing values and data heterogeneity must be considered when designing a machine-learning architecture to treat EHR data.

\begin{figure*}[t]
    \begin{center}
    \centerline{\includegraphics[width=\textwidth]{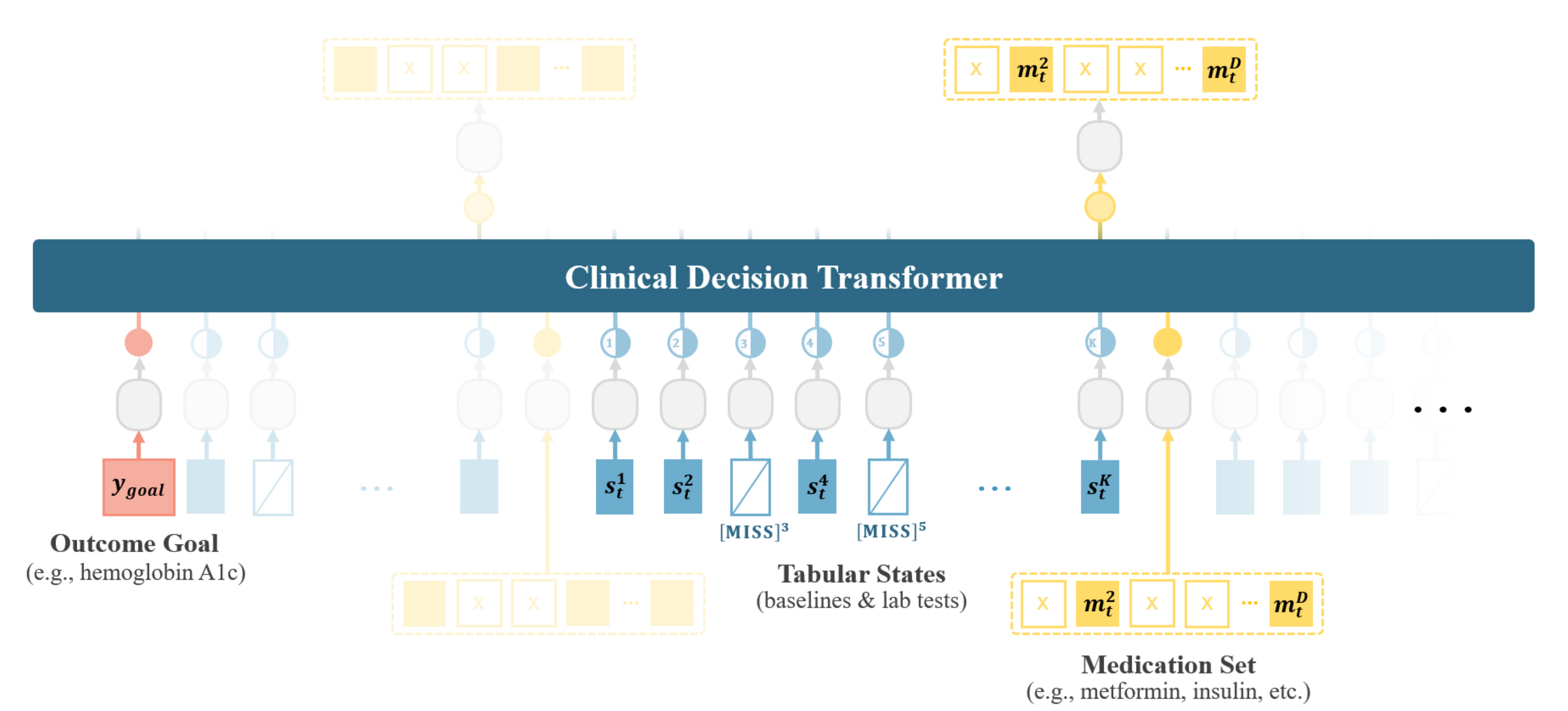}}
    \caption{Overview structure of the Clinical Decision Transformer. A pair of the desired future goal state and the clinical-treatment history consisting of tabular states and medication sets are embedded separately and fed into the GPT architecture. The unmeasured missing values in the states are represented by trainable special tokens of $\left [ \textrm{MISS} \right ]^k$. By manipulating the goal prompts, the model recommends intended medications conditioned on the goal state and treatment history.}
    \label{figure1}
    \end{center}
\vskip -0.2in
\end{figure*}

Driven by the recent maturation of the pretraining paradigm on broad data, foundation models \cite{bommasani2021opportunities} have been demonstrated to offer solutions to various tasks requiring high-level context awareness. GPT-3 \cite{brown2020language} can generate output sentences through “in-context learning” when example prompts consisting of input-output pairs are provided. As a commercialized pair-programming agent, the fine-tuned GPT-3 in public GitHub repositories, Copilot \cite{chen2021evaluating}, recommends the next code lines according to the codes and comments typed prior. Furthermore, Decision Transformer \cite{chen2021decision} and Trajectory Transformer \cite{janner2021offline}, both based on the GPT architecture \cite{radford2018improving}, verified that the decision-making task could be cast as a sequence-modeling framework by demonstrating long-term optimized action selections conditioned on past interactions. Furthermore, from the same framework, Multi-Game Decision Transformer \cite{lee2022multi} and Gato \cite{reed2022generalist} revealed the generalist agents that could adapt to distinct tasks based on the input context. 

Similarly, in the medical domain, emerging studies have applied foundation models to address the contextualization problem for EHR data. BEHRT \cite{li2020behrt} and Med-BERT \cite{rasmy2021med}, rooted in the BERT \cite{devlin2018bert} architecture, receive a series of classification codes for diagnosis over several hospital visitations. Subsequently, these models were employed to predict future disease diagnoses by conjugating pre-trained embeddings of the input code histories. Unlike both models that predict disease prognosis, G-BERT \cite{shang2019pre} predicts the next treatments from the previous visitations. In this process, it generates ontology embeddings of the diagnoses and treatments through a graph neural network,  after which BERT projects them to visitation embeddings for each hospital admission. However, the above three studies do not utilize laboratory test results from EHR systems; instead, they solely use diagnosis codes, which are less informative in reflecting individual heterogeneity. Moreover, due to their low necessity, several codes are not inputted into the EHR systems, even though they are diagnosed or treated.

In this work, we designed Clinical Decision Transformer (CDT), a GPT-based sequential medication recommender system that directly utilizes the laboratory test data with missing values and heterogeneity. Notably, G-BERT also aims to recommend medication; however, it is not structurally possible for G-BERT to achieve the intended direction as it is based on behavior cloning of the prescriptions in the EHR data. In contrast, the proposed CDT can recommend intended medications according to a preassigned goal, which acts like a “prompt” that is extensively used in natural language foundation models. 

CDT generates medications conditioned on (1) a desired clinical goal state and (2) a sequential history from individual EHR data. In our experiment, we trained CDT on an extracted EHR dataset of patients with diabetes. Treatment effects of the recommended medications depending on the assigned goals, were evaluated based on the HbA1c (A1c) values. To estimate counterfactual clinical outcomes, as we do not have access to ground-truth A1c, we trained the additional evaluation model Counterfactual Recurrent Network \cite{bica2020estimating}. To the best of our knowledge, this is the first study to propose a clinical recommender system based on goal prompting with foundation models. Our contributions are as follows:

\begin{enumerate}
    \item We designed the model to recommend medications conditioned on the clinical goal prompt that we want patients to achieve as long-term disease management.
    \item We applied column embeddings, special tokens $\left [ \textrm{MISS} \right ]$, and admission-wise attention masks to the GPT architecture to capture context from tabular data with missing values.
    \item We evaluated the treatment effects of the recommended medications by training an auxiliary model for counterfactual estimation.
\end{enumerate}

\section{Related Work}

\textbf{Decision-making via sequence modeling.} Inspired by the remarkable achievements of foundation models in natural language and computer-vision tasks, a concept that considers decision-making as a sequence-modeling problem has emerged. Decision Transformer \cite{chen2021decision} maps received past trajectories to subsequent action distribution, and it is possible to produce improved actions by conditioning high returns in the input trajectory. Similarly, the Trajectory Transformer \cite{janner2021offline} maps input sequences to joint distributions of state, reward, and action. Notably, the Trajectory Transformer demonstrates that the concept of decision-making as a sequence-modeling problem is also applicable to goal-conditioned trajectory generation. By simply prepending the last state to the initial token of the input sequence, the model produces the most possible future trajectory to reach that state. For our medication recommender system, we adopt two approaches: modeling action distributions and generating sequences through goal conditioning.

\textbf{Multivariate-time series with missing values.} A high proportion of missing data is a frequent problem in clinical data utilization. A simple and plausible approach is to omit the missing data \cite{schafer2002missing}; however, this approach can cause selection bias if the missing rate is high or the missingness is associated with meaningful information. Another approach is to conduct imputation, such as smoothing, spline, and interpolation. As the imputation step separates from the training step for tasks (e.g., prediction), models trained with imputed data do not intentionally consider the missing context. Moreover, inefficiency and bias can arise from these two-step processes if the data volume increases. In this study, for end-to-end training, we utilized column embeddings and special tokens for the missing values.

\textbf{Counterfactual estimation using deep learning.} In evaluating the treatment effects using observed data, as an emulated randomized clinical trial, counterfactual estimation is essential, and so is selection bias reduction. Compared with traditional methods, deep learning-based approaches for balanced representation have shown improved performance with the presence of time-dependent confounding. For example, the Counterfactual Recurrent Network \cite{bica2020estimating} uses a gradient reversal layer in the training process, and the Causal Transformer \cite{melnychuk2022causal} adopts the domain confusion loss with three separate subnetworks. To estimate the counterfactual outcomes according to the medications recommended by CDT, which are different from factual prescriptions, we implemented the encoder architecture from the Counterfactual Recurrent Networks.

\section{Problem Formulation}
\label{section3}

An EHR dataset, $\mathcal{D}=\left\{ \left ( s^{(i)}_t,a^{(i)}_t,y^{(i)}_{t+1},  \right )^{T^{(i)}}_{t=1} \right\}^N_{i=1}$, consists of $N$ patients’ clinical state and medication histories. Each sequence $\left(i\right)$ represents consecutive health records of a patient $\left(i\right)$ for $T^{\left(i\right)}$ hospital admissions. For each admission, we observe clinical states $\mathbf{S}_t^{\left(i\right)}$, sets of prescribable medications $\mathbf{A}_t^{\left(i\right)}$, and treatment outcomes $\mathbf{Y}_{t+1}^{\left(i\right)}$. The clinical states correspond to the tabular data: $\mathbf{S}_t=\left\{\mathbf{B}_t^1,\cdots,\mathbf{B}_t^J,\mathbf{L}_t^1,\cdots,\mathbf{L}_t^G\right\}$, where $\mathbf{B}^j$ is the $j^{th}$ column value of the baseline covariates (e.g., gender, age, and presence of diagnosis for several diseases), and $\mathbf{L}^g$ is the $g^{th}$ column value of the laboratory test results (e.g., blood glucose, blood pressure, or certain protein concentrations in the blood), measured on each admission. For simplicity, let the notation of the tabular data denote baseline columns and laboratory test columns in an integrated manner $\left\{\mathbf{S}_t^1,\cdots,\mathbf{S}_t^K\right\}$, where $K=J+G$. The medication set, $\mathbf{A}_t\in\mathcal{P}\left(\left\{m^1,\ \cdots,m^D\right\}\right)=\left\{\emptyset,\ \left\{m_1\right\},\ \left\{m_1,\ m_2\right\},\cdots\right\}$, is a power set of D prescribable medications (e.g., metformin, insulin, or sulfonylurea). The treatment outcome, $\mathbf{Y}_t\in\mathbb{R}$ is the range of a specific covariate (e.g., A1c) that clinicians focus on to manage. Here we assume that the laboratory test results contain missing values in some columns, and so do the treatment outcomes. Furthermore, in this work, the treatment outcome is set as a clinical variable, which is better when the value is relatively low (e.g., tumor volume or A1c excluding extremely low values). For a simplified explanation, we omit the patient index $\left(i\right)$ unless required.

Let ${\mathbf{\bar{H}}}_{t}^{t^\prime}=\left\{{\mathbf{\bar{S}}}_{t}^{t^\prime},{\mathbf{\bar{A}}}_{t}^{t^\prime}\right\}$, where ${\mathbf{\bar{S}}}_{t}^{t^\prime}=(\mathbf{S}_{t},\cdots,\mathbf{S}_{t^\prime})$ and ${\mathbf{\bar{A}}}_{t}^{t^\prime}=\left(\mathbf{A}_{t},\cdots,\mathbf{A}_{t^\prime}\right)$, represent the partial history of patient admissions. For goal-conditioned sequence modeling, each patient’s sequence is divided into partial sequences at admission points, followed by the lowest treatment outcome up to the last admission:
\begin{equation}
    \mathbf{Y}_{goal}=\min{({\mathbf{\bar{Y}}}_{t+1}^T)},   
\end{equation}
where ${\mathbf{\bar{Y}}}_{t+1}^T=(\mathbf{Y}_{t+1},\cdots,\mathbf{Y}_T)$ and $t$ is an admission point with the previous lowest outcome.  
These lowest outcomes are set to be goals for each partial history. Thus, the goal-conditioned partial history for training is as follows:
\begin{equation}
    \left\{\mathbf{Y}_{goal},\ {\mathbf{\bar{H}}}_{t}^{t^\prime}\right\}=\left\{\mathbf{Y}_{{t^\prime}+1},\ {\mathbf{\bar{S}}}_{t}^{t^\prime},{\mathbf{\bar{A}}}_{t}^{t^\prime}\right\}.            
\end{equation}

\begin{figure}[t]
\vskip -0.1in
\begin{center}
\centerline{\includegraphics[width=\columnwidth]{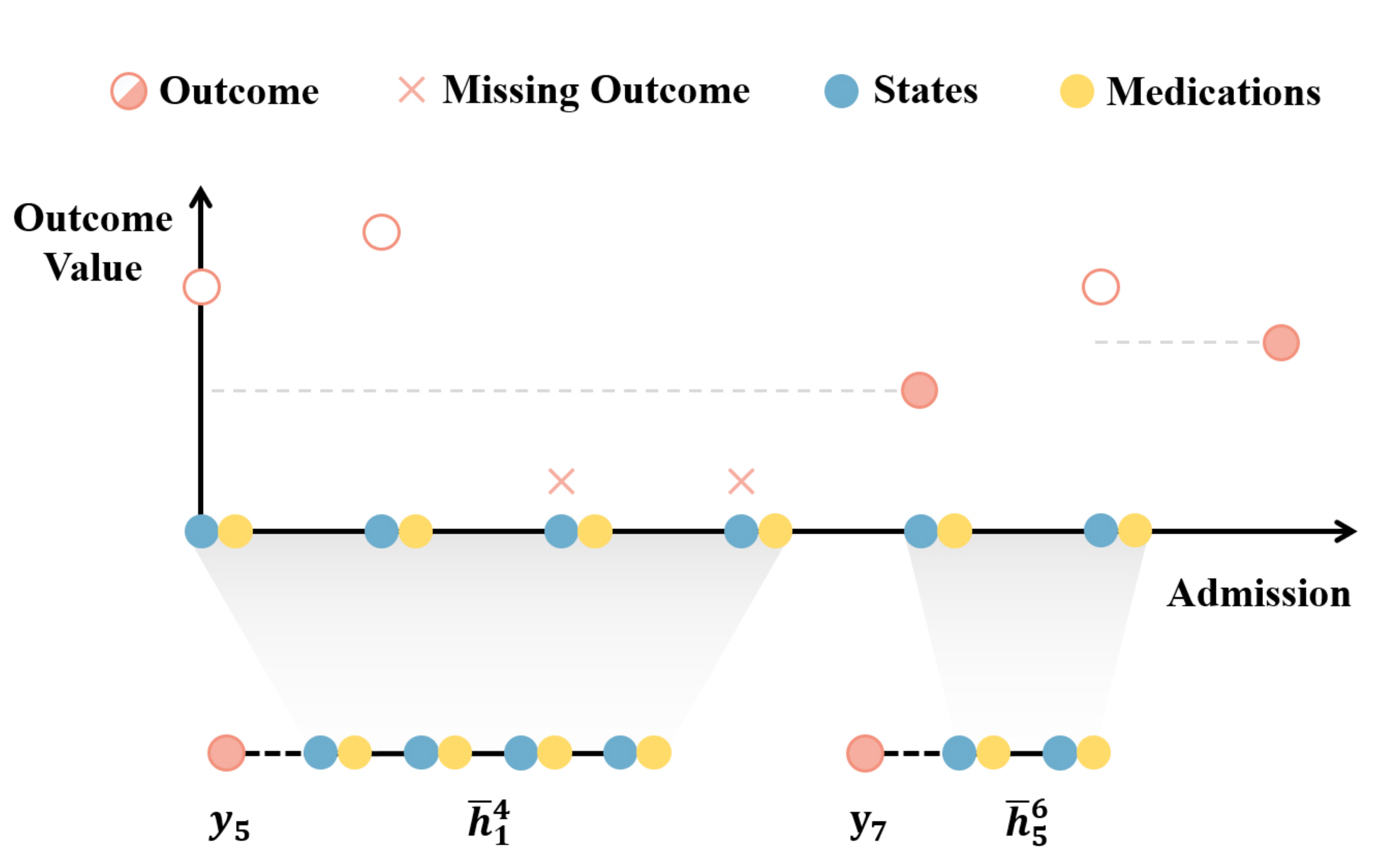}}
\caption{Goal-conditioned sequencing process. A clinical history consisting of several times of hospital admissions is sliced at the lowest treatment outcome (if lower outcome values are desired), and these lowest outcomes are prepended to each subsequence.}
\label{figure2}
\end{center}
\vskip -0.4in
\end{figure}

\cref{figure2} illustrates this goal-conditioned sequencing process. As $y_5$ is the lowest outcome up to the last admission, the whole sequence with seven admissions is sliced at the fourth admission. Subsequently, the second  lowest outcome is $y_7$, and the whole sequence is divided into two subsequences, ${\bar{h}}_1^4$ and ${\bar{h}}_5^6$. For goal-conditioned sequencing, the two lowest outcomes, represented as filled red circles, are respectively prepended to the previous subsequences as the outcome goals. Note that the missing outcomes notated with red crosses are not considered target outcomes. For simplicity in the notation of the preprocessed subsequences, the first admission point is rephrased as 1. In this study, we proposed an intended clinical recommendation through goal prompting and intended to estimate medications conditioned on the assigned outcome goal, previous sequence, and current states:
\begin{equation}
    \mathbb{E}(\mathbf{A}_t|\mathbf{Y}_{goal},\ {\mathbf{\bar{H}}}_1^{t-1},\mathbf{S}_t).
\end{equation}

\section{Clinical Decision Transformer}

Herein, we describe the architecture of the proposed model through which it generates clinical recommendations according to assigned treatment goals and individual EHR histories with missing values. An overview of the architecture is shown in \cref{figure1}.

\subsection{Model architecture}

Similar to the Decision Transformer and the Trajectory Transformer, CDT uses the causal GPT architecture for a decision-making problem. Notably, both studies optimize the action by modulating the return-to-go to be high as a reinforcement learning framework. However, since the missing rate of covariates at each admission is generally high in the EHR data, it is not easy to calculate the return-to-go for those sparse features as an optimization objective. Instead, CDT uses a goal-conditioned sequence generation framework, which requires only one goal prompt per sequence. Thus, it is rarely affected by the missing rate of intermediate covariates. The goal-conditioned sequence generation aims to model the conditional probability of the past sequence given the desired future treatment outcome. In the CDT, we set the lowest outcome up to the last admission as an outcome goal, $y_{goal}$, by the sequencing process explained in \cref{section3}. Thereafter, a preprocessed goal-conditioned subsequence $\left (y_{goal},\bar{h}^T_1 \right )$ with $T$ admissions is rephrased into a conditional input sequence and the medication set to be estimated:
\begin{equation}
    \left\{ \left ( y_{goal},\bar{h}^{t-1}_{1},s_{t},a_{t} \right ) \right\},
    \quad \textrm{for} \ t=1,\cdots,T.
\end{equation}
With this input-output pair, the objective we intend to maximize during training is:
\begin{equation}
    \mathcal{L} \left (\theta \right) = \displaystyle\sum_{t=1}^{T}log{P_{\theta}\left(a_{t}\mid y_{goal},\bar{h}^{t-1}_{1},s_{t} \right)}, 
\end{equation}
denoting the parameters of the CDT as $\theta$.

\subsection{Column embedding}

When dealing with sequential EHR data in practice,  we need to consider time-series tabular data with a high missing rate. We adopted column embedding \cite{huang2020tabtransformer, gorishniy2021revisiting} for the CDT to distinguish each column from others. Let $c_{\phi_k}$ be the column identifier embedding and $v_{\phi_{k,t}}$ be the value embedding of the $k^{th}$ column at admission t. For baseline columns that were categorical features, entire classes were encoded to integers and comprehensively embedded. For laboratory test columns that were continuous features, values were normalized column-wise and embedded through separated embedding layers. The missing occurrence only corresponded to the laboratory tests. We set special tokens, $\left [ \textrm{MISS} \right ]^k$, to represent the omissions in each laboratory test column. These special tokens were independent of each other and separately embedded as value embeddings, $v_{\phi_k}$, when the histories contained missing values. We derived the column embeddings of clinical states by concatenating the corresponding column identifier embeddings and value embeddings:
\begin{equation}
    e_{\phi}\left ( s^k_t \right ) =
    \left [ c_{\phi_k},v_{\phi_{k,t}}\right ],
\end{equation}
where $e_\phi$ represents the conprehensive embedding layers for $t=1,\cdots,T$ and $k=1,\cdots,K$.

The power set of prescribable medications was categorized, followed by the embedding process. For the prepended outcome goals, as they were generally numerical features, we discretized the range and applied categorized embeddings. As in the Decision Transformer, CDT did not use per-token positional embedding. Instead, we added identical admission embedding to the clinical state tokens and the medication token in the same admission.

\begin{figure}[t]
    \begin{center}
    \centerline{\includegraphics[width=\columnwidth]{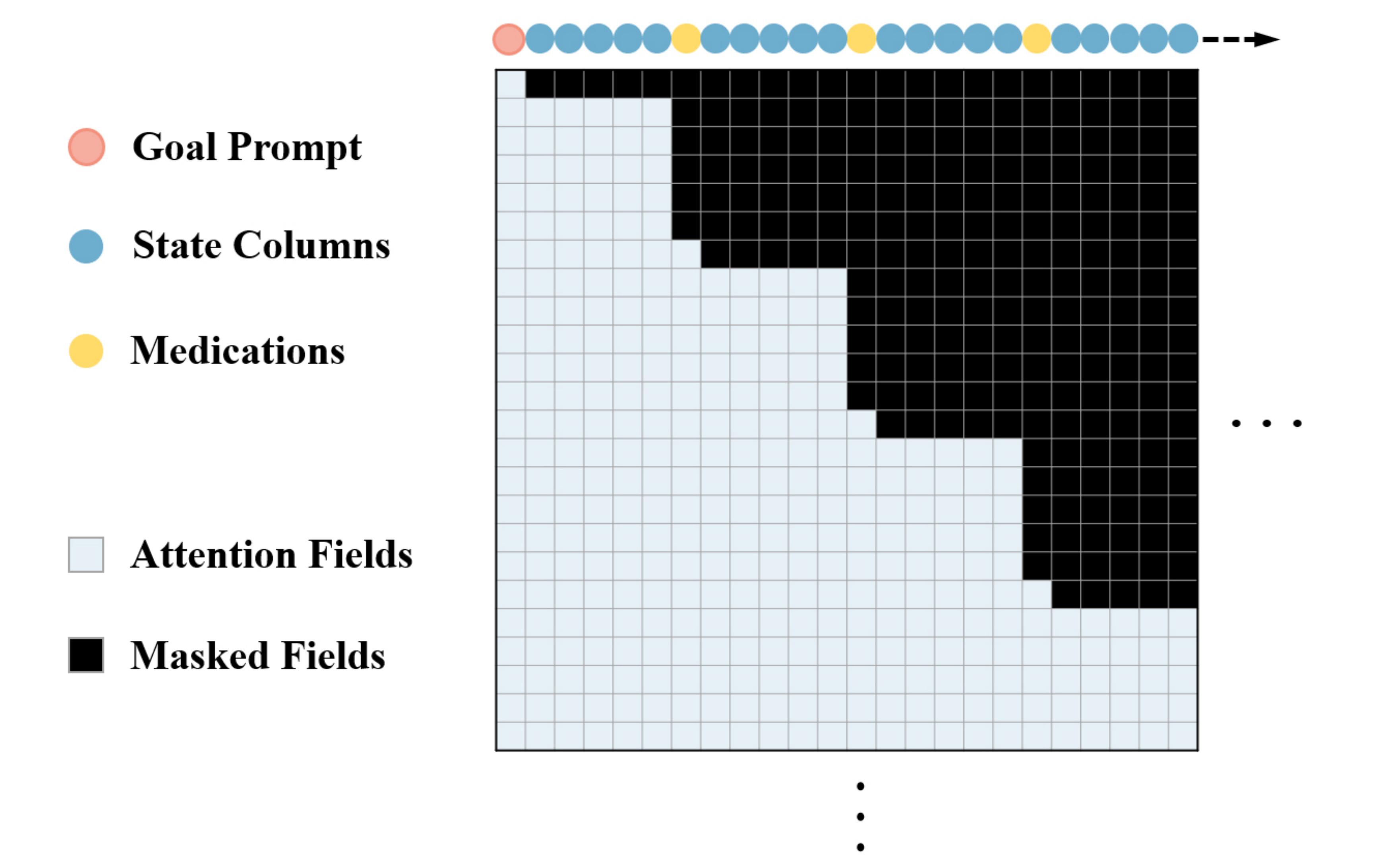}}
    \caption{Admission-wise attention mask. Modified from the causal mask of the GPT, attention field for all state tokens are extended until the last state tokens of the same admission.}
    \label{figure3}
    \end{center}
    \vskip -0.3in
\end{figure}

\subsection{Admission-wise attention mask}

For the contextual embedding, we designed an admission-wise attention mask, different from the token-wise causal mask of the GPT. \cref{figure3} shows a simplified illustration of our admission-wise attention mask, where the clinical states have five columns. We extended the attention field for all state tokens of the same admission. Through the forward pass of the GPT layers, this attention field allows each clinical state token to be contextualized over not only inter-admissions but also intra-admissions. In experiments, we examined how the admission-wise attention contextualized missing values and how effective it is in the overall architecture.

\section{Experiments}

\subsection{Diabetes EHR dataset}

We extracted cohort samples from an EHR system. We screened 4788 patients diagnosed with diabetes who entered a hospital between January 2015 and July 2021. After the goal-conditioned sequencing, the total batch size of subsequences was 9406, and the maximum admission number was 34 times. The dataset had 14 and 18 columns for the baseline and laboratory tests, respectively. Notably, the average missing rate of the laboratory test columns was high at 44.39\% (median: 22.37\%). There were 12 prescribable medications, and of its power sets, 441 subset classes were used for training, excluding the combination where the case was not in the dataset. As A1c is the primary measure for diabetes management, we used the A1c as an outcome for goal prompting. We categorized the A1c range from 4.0\% to below 18.0\% with a 0.1\% increment. The missing rate of the A1c feature was 37.30\%. See \cref{appendixD} for details of the dataset. 

\subsection{Training}

We implemented the CDT based on the GPT-2 of the Huggingface Transformer library \cite{wolf2019huggingface}. Conditioned on an input sequence consisting of a prepended goal prompt, previous admission history, and current clinical states, the CDT was trained to predict the current medication set prescribed by clinicians with cross-entropy loss. To prevent the attention operation from overflowing, we adopted QKNorm \cite{henry2020query}, L2 normalization of queries and keys before the dot product. We split the subsequences into training, validation, and testing datasets in an 8:1:1 ratio. The hyperparameters of the CDT were optimized on the validation set, as explained in \cref{appendixB}. In the experiments, we evaluated the treatment effect of the medications recommended by the CDT according to goal prompts. For this purpose, we adopted an encoder part of the Counterfactual Recurrent Network \cite{bica2020estimating} as an evaluation model. We trained the model to estimate the counterfactual A1c of the one-step subsequent admission according to the recommended medications that could differ from the factual prescriptions. The details of the evaluation model, candidate architectures, and their training process are provided in \cref{appendixA}. As a baseline for our prompting-based intended recommendation, we trained a behavior-cloning (BC) model, which imitates the factual prescriptions of clinicians when patients’ histories are given. This BC model was built upon long short-term memory (LSTM) network and trained with a supervised loss.

\subsection{Recommendation through a goal prompting}

\begin{figure}[t]
    \begin{center}
    \centerline{\includegraphics[width=\columnwidth]{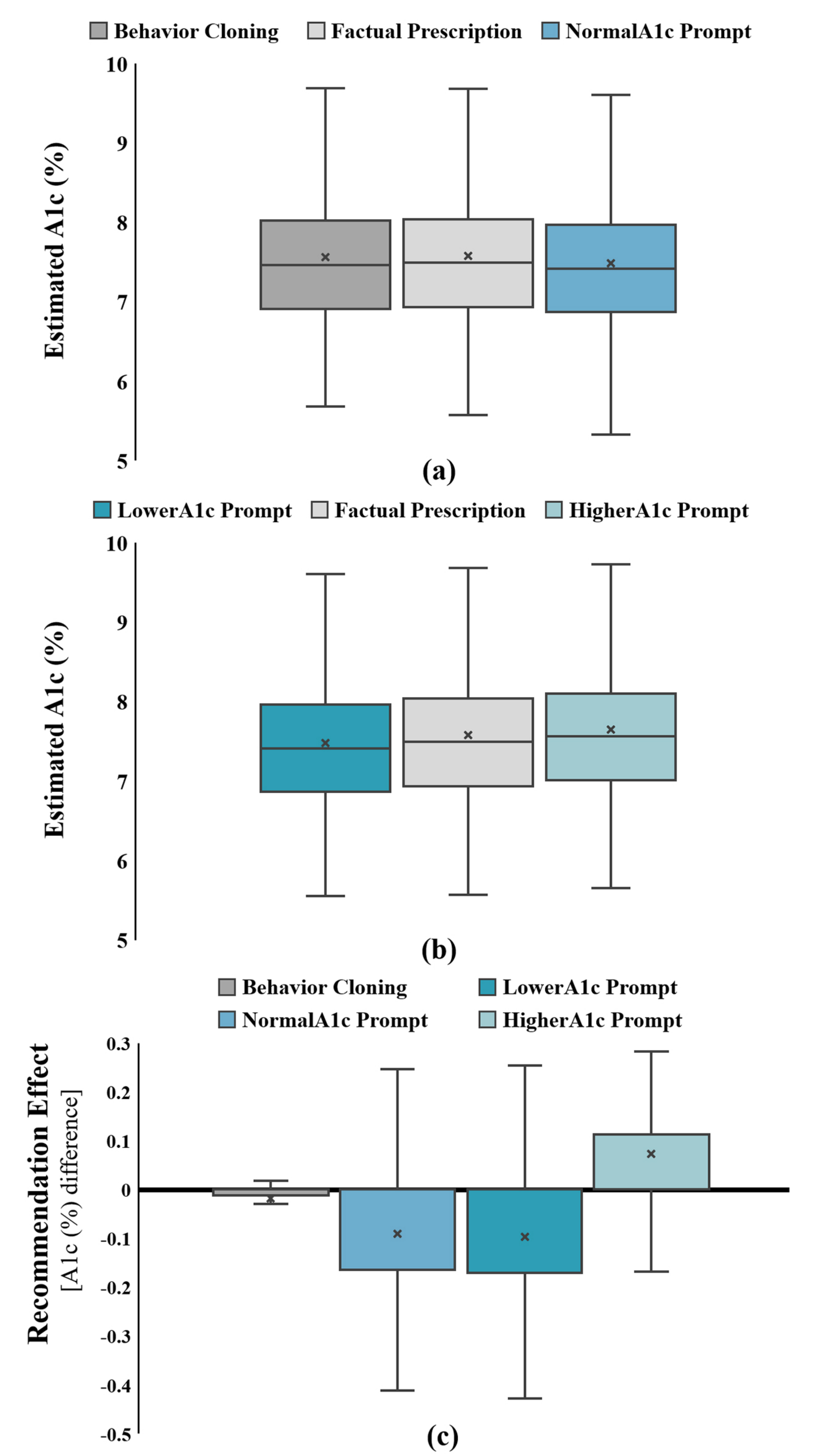}}
    \caption{Distributions of estimated A1c values and recommendation effects according to the goal prompt ranges.}
    \label{figure4}
    \end{center}
    \vskip -0.5in
\end{figure}

We evaluated whether the effects of the medications recommended by the CDT conditioned on the goal prompt were as intended. We manipulated the prepended goals of each goal-conditioned subsequence in three prompt ranges: NormalA1c, LowerA1c, and HigherA1c. The NormalA1c prompt represents the range of A1c that is considered normal by the American Diabetes Association, which is 4.0\%–-5.6\%. The LowerA1c prompt represents an A1c range between 4.0\% and below the pre-existing goals in the subsequences. Similarly, the HigherA1c prompt corresponds to an A1c range between above the pre-existing goal and below 18\%. According to the goal-conditioned sequencing process, these pre-existing goals are actual treatment outcomes that are achieved in the future after factual prescriptions. We sequentially conditioned all A1c tokens in each prompt range and selected a medication set with the highest probability for a recommendation. We estimated the A1c outcomes of the  next admission using the trained evaluation model for the selected medication sets and the factual prescriptions. The A1c-lowering effect of recommendation was measured as a difference in each estimated next A1c value by factual prescription and recommended medication set. The results are shown in \cref{figure4} and \cref{table1}. 

\begin{table}[t]
    \caption{Average recommendation effects according to the goal prompt ranges.}
    \label{table1}
    \begin{center}
    \begin{small}
    \renewcommand{\arraystretch}{1.2}
    \begin{tabularx}{\columnwidth}{Xc}
    \Xhline{3\arrayrulewidth}
    \textbf{Prompt Range} & \textbf{A1c(\%) Difference}\\
    \hline\hline
    Behavior Cloning    & \textminus0.0181\\
    \midrule
    NormalA1c (4\% -- 5.6\%) & \textminus0.0914\\
    LowerA1c (4\% -- pre-exiting goal)   & \textminus0.0968\\
    HigherA1c (pre-exiting goal -- 18\%)    & +0.0727\\
    \Xhline{3\arrayrulewidth}
    \end{tabularx}
    \end{small}
    \end{center}
    \vskip -0.1in
\end{table}

\cref{figure4} (a) shows that the distribution of A1c by behavior cloning is similar to the distribution by factual prescription. On the other hand, with the NormalA1c prompt, the distribution of A1c exhibited a downshift to the normal range as intended. In \cref{figure4} (b), with relative goal promptings, the A1c distributions by the LowerA1c and HigherA1c prompts were downshifted and upshifted, respectively, compared to the distribution by factual prescription. To highlight the recommendation effect, \cref{figure4} (c) shows how much each regime lowers A1c compared to the factual prescription. 

Unlike the marginal improvement in behavior cloning, the prompt-based medications caused a relatively significant treatment effect according to the conditioned goal. From the comparison results, we can conclude that the clinical recommendations through the goal prompting had the intended treatment effects. 

Note that these comparisons covered the treatment effects of the very next admission, not the long-term goal achievement. Although we did not design the evaluation model for an autoregressive inference purpose to avoid unreliable bias from the high missing rate of the EHR data, autoregressive counterfactual estimation with missing values could be explored in future work.

\subsection{Recommendation analysis}

We investigated the performance of the CDT and the BC model compared with the factual prescription by clinicians. As the outputs of the CDT were classes of mediation sets, we factorized the recommended set into medication elements and evaluated the item-wise performance in the analysis. We selected two metrics for practical clinical recommendation: (1) the recall for examining how the model pre-assists clinicians before they make decisions on-site and (2) the false-positive rate (FPR) for less interfering with clinicians’ decisions, avoiding misuse of medications.
\begin{equation}
    Recall=\small{\frac{True\ Positive}{True\ Positive\ +\ False\ Negative}}
\end{equation}
\begin{equation}
    FPR=\small{\frac{False\ Positive}{False\ Positive\ +\ True\ Negative}}
\end{equation}

We compared the output according to the input admission length to analyze the utilization of inter-admission and inter-admission information.

\begin{figure}[t]
    \begin{center}
    \centerline{\includegraphics[width=\columnwidth]{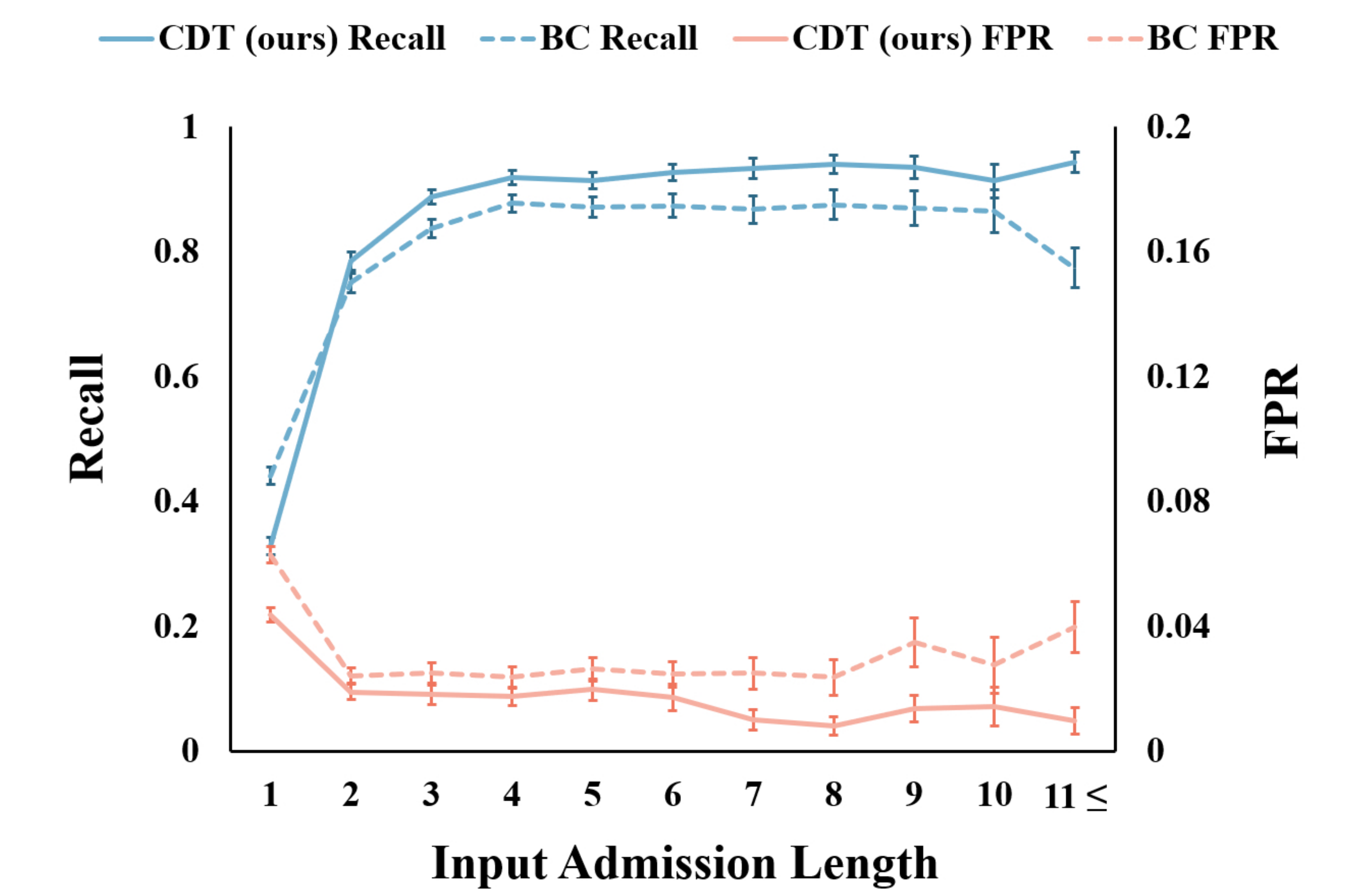}}
    \caption{Recommendation performance curves of the CDT and behavior cloning (BC). Along the length of the input admission, recalls and false positive rates are compared as item-wise evaluation metrics.}
    \label{figure5}
    \end{center}
    \vskip -0.4in
\end{figure}

\begin{table}[h]
\caption{Recommendation metrics according to the length of input admissions.}
\label{table2}
\begin{center}
\begin{small}
\renewcommand{\arraystretch}{1.1}
\begin{tabularx}{\columnwidth}{XXXXX}
\Xhline{3\arrayrulewidth}
                                     &              & \multicolumn{3}{c}{\textbf{Input Admission Length}} \\ \cline{3-5} 
                                     &              & Cold Start          & 2$\leq$             & 11$\leq$            \\ \hline\hline
\multirow{2}{*}{\textbf{Recall(\%)}} & \textbf{CDT}(ours) & 32.9                & \textbf{88.5}          & \textbf{94.4}          \\
                                     & \textbf{BC}  & \textbf{44.1}                & 83.2          & 77.4          \\ \hline
\multirow{2}{*}{\textbf{FPR(\%)}}    & \textbf{CDT}(ours) & \textbf{4.47}                & \textbf{1.65}          & \textbf{0.97}          \\
                                     & \textbf{BC}  & 6.29                & 2.59          & 3.97 \\      
\Xhline{3\arrayrulewidth}
\end{tabularx}
\end{small}
\end{center}
\vskip -0.1in
\end{table}

\textbf{Results.} In this analysis, cold start corresponds to the case where the input admission length is one, and it includes both the first and non-first hospital visits according to the goal-conditioned sequencing process. \cref{figure5} shows the recall and FPR curves according to each input length, where 11 is the top 15\% percentile admission length. We observed that recall of the CDT was approximately 33\% at the cold start and drastically increased to 94\% as the input admission length increased. In contrast, the FPR was around 4\% at the cold start case and decreased to 1\% when more input admissions were given, finding that the CDT utilized information from previous admissions. Although the recall of the BC was higher than that of the CDT at the cold start, its FPR was also higher, meaning the BC model tended to abuse recommendations. For inputs with longer lengths, the CDT recommended medications better than the BC, considering higher recall and lower FPR. \cref{table2} describes the detailed metrics. These results support the potential of the CDT as a clinical recommender system, which should be conservative.

\begin{figure}[t]
    \begin{center}
    \centerline{\includegraphics[width=\columnwidth]{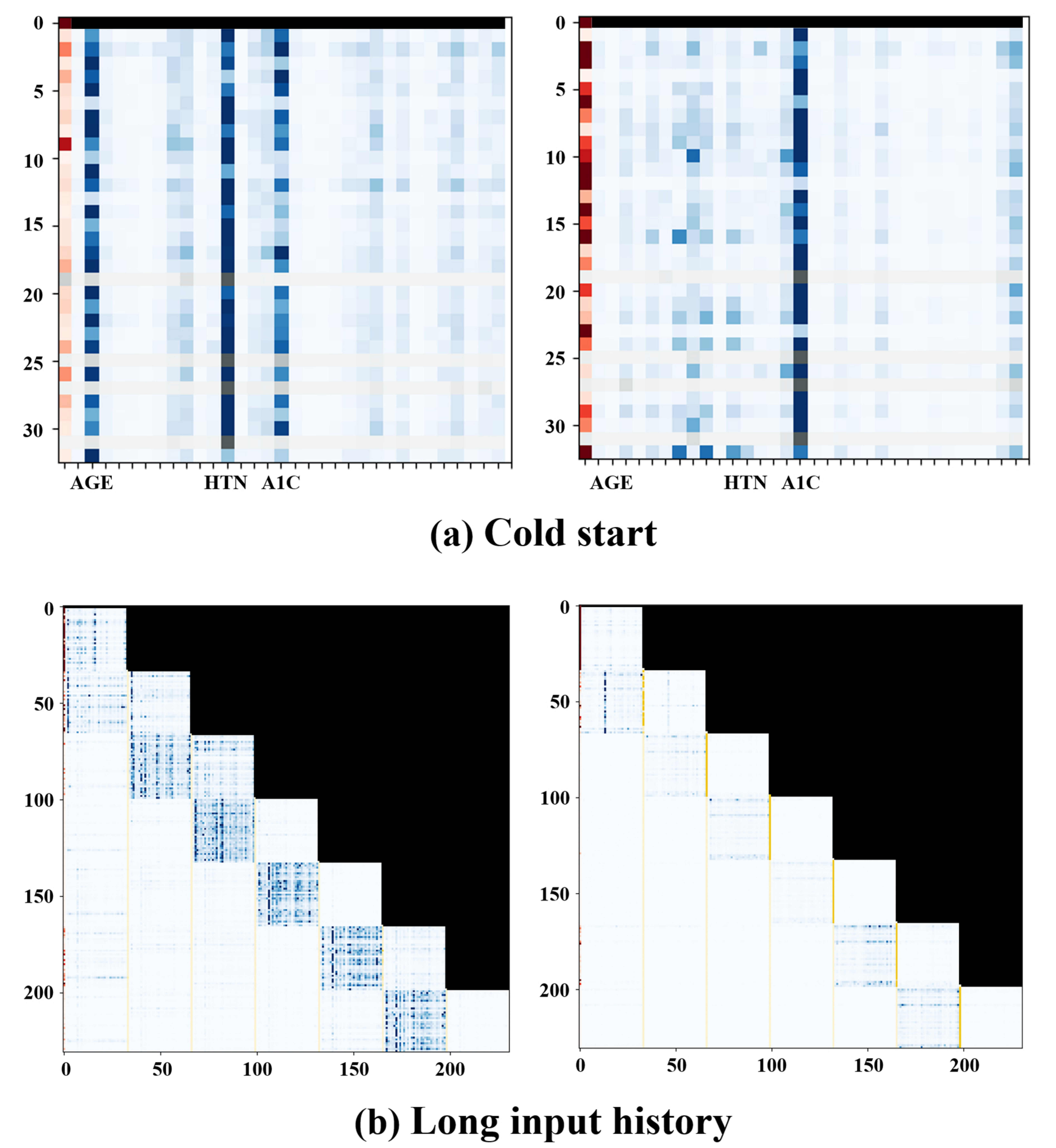}}
    \caption{Representative attention patterns observed in inference of the CDT. (a) Cold start case; age, hypertension (HTN), and A1c were priorly utilized in contextualizing each state, even in attention weights for missing values colored in gray scale. (b) Long input history; two kinds of states-focused and medication-focused attention patterns were observed. In both patterns, recent information was more highly weighted than previous ones.}
    \label{figure6}
    \end{center}
    \vskip -0.4in
\end{figure}

\textbf{Attention patterns.} We visualized representative cases of attention patterns to closely analyze how the CDT recommended medications and contextually embedded sequential EHR data. \cref{figure6} (a) shows the attention patterns of the first layer observed in the cold start case. The leftmost red column corresponds to a goal, and the other blue columns represent the clinical states of the baseline and laboratory tests. The gray-colored rows are attention weights regarding the special tokens $\left [ \textrm{MISS} \right ]$ for missing values. In the first heads, the CDT contextualized each state feature primarily utilizing age, hypertension, and A1c. It is noteworthy that age and A1c are major features that clinicians pay attention to when prescribing medications to patients with diabetes in practice. We could observe that missing values were contextualized with attention to other diagnoses and laboratory test information from the same admission. In the second head, attention weights were high on the goal and A1c tokens. When the given length of previous admissions became longer, the attention patterns were characterized by a combination of two different head patterns: state-focused head and medication-focused head. \cref{figure6} (b) shows the multi-head attention patterns of the same layer, where the yellow columns correspond to previous medication sets. In the first head, the model highly attended to the states of the immediate last admission, focusing on recent information rather than old ones. In the second head, the model attempted to utilize the prescription information of the last admission for recommending current medications.

\begin{figure}[t]
    \begin{center}
    \centerline{\includegraphics[width=\columnwidth]{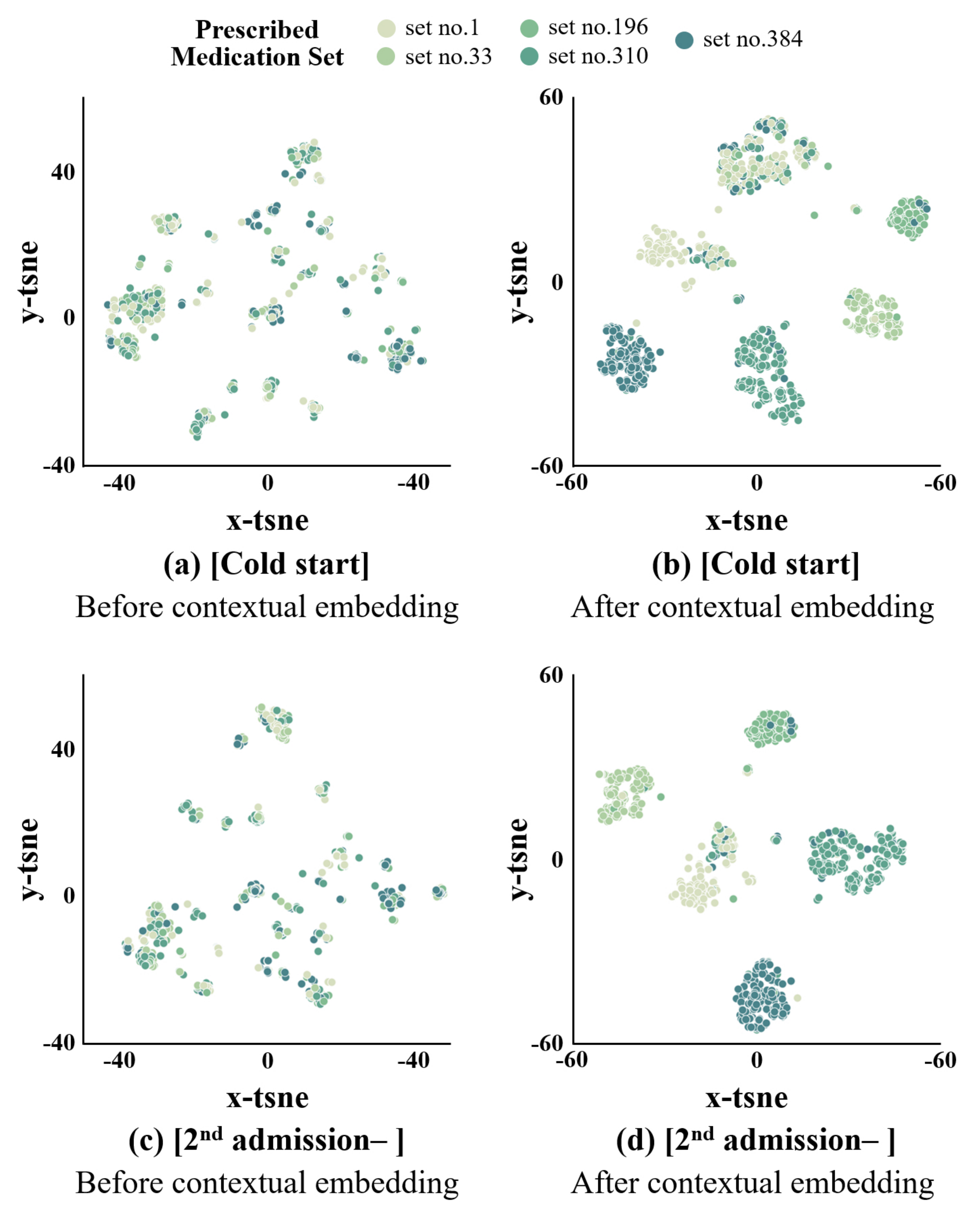}}
    \caption{T-SNE embeddings of the admissions before and after contextual embedding.}
    \label{figure7}
    \vskip -0.3in
    \end{center}
\end{figure}

\textbf{Contextual embedding.} To explicitly evaluate whether the CDT conducts contextual embedding, in \cref{figure7}, we illustrate each admission point in the embedding space before and after passing the GPT layers. Each color represents a set of factual prescriptions. Input features and embedded output vectors corresponding to each admission are projected onto two dimensions through the t-SNE method. As the plots show, we observe that the output embeddings are clustered with respect to prescriptions, while the input admissions are entangled before contextualized.

Considering these recommendation analyses, we can comprehensively conclude that the model tends to utilize recent inter-admission information to contextualize the clinical states of the current admission, while when previous information was insufficient, intra-admission contextualization became more active. 

\subsection{Ablation study}

The CDT architecture is characterized by column embedding and the admission-wise attention mask. We ablated these architectures from CDT and compared effects on recommendation performance. As a single statistic metric, we used the Youden index \cite{youden1950index}, defined as a gap between recall and FPR and widely used for evaluating the performance of clinical tests and models \cite{schisterman2005optimal, powers2020evaluation}.
\vspace{-0.05in}
\begin{equation}
    Youden\ index=Recall-FPR.  
\end{equation}
\vskip -0.1in
\cref{table3} shows the relative changes in the metric compared with the result of CDT. Without the column embedding, recommendation performance decreased in all lengths of input admissions because of the difficulty in representing tabular data. On the other hand, the admission-wise attention mask was effective only in the cold start case. This result matches the attention patterns illustrated in \cref{figure6} (b), where the attention weights for the previous admissions were higher than those for the current admission. 

\begin{table}[t]
\caption{Ablation study for CDT architecture. Compared to the Youden index of CDT, differences were calculated. (CE: column embedding, AwMask: admission-wise attention mask)}
\label{table3}
\begin{center}
\begin{small}
\renewcommand{\arraystretch}{1.3}
\begin{tabularx}{\columnwidth}{Xcc}
\Xhline{3\arrayrulewidth}
                                            & \multicolumn{2}{c}{\textbf{Youden Index on Input Length}} \\ \cline{2-3} 
                                            & Cold Start                    & 2$\leq$                         \\ \hline
\textbf{CDT}                                & 0.2849                        & 0.8688                    \\ \hline
\textbf{w/o CE}              & \textminus 0.0170                       & \textminus0.0132                   \\
\textbf{w/o AwMask} & \textminus0.0051                       & 0.000                    \\
\textbf{w/o CE \& AwMask}                           & \textminus0.0322                       & -0.0111                   \\
\Xhline{3\arrayrulewidth}
\end{tabularx}
\end{small}
\end{center}
\vskip -0.2in
\end{table}

\section{Conclusion}

In this paper, we propose a Clinical Decision Transformer, a clinical recommender system that suggests medications in intended directions according to conditioned goal prompts. We extracted a dataset of 4788 patients with diabetes from the EHR systems and conducted goal-conditioned sequencing to generate goal-sequence pairs. We trained a GPT-based foundation model to recommend a sequence of medication sets required to reach a prompted A1c goal when a treatment history and current states were given. It is necessary to contextualize inter-admission and intra-admission information to fully utilize large-scale EHR data characterized by heterogeneity and high missing rates. For this purpose, we applied the admission-wise attention mask to the GPT architecture for contextual embedding and adopted the column embedding to enable the model to capture structural information of tabular data. In the experiment, we observed that the CDT achieved the intended treatment effects according to given goal prompt ranges, such as NormalA1c, LowerA1c, and HigherA1c. Furthermore, the CDT recommended medications by embedding the context through inter-admission and intra-admission with characteristic attention patterns. With the proposed approach of considering clinical recommender systems as a goal-prompting problem through foundation models, the scalable utilization of large-scale and disease-wide computerized EHR systems could be a socially meaningful future research direction.

\section*{Fundings}

This work was supported by the National Research Foundation of Korea (NRF) grant funded by the Korea government (MSIT) (No. 2020R1A2C2005385), the Basic Science Research Program through the NRF funded by the Ministry of Education (No. 2020R1A6A1A03047902), the Bio \& Medical Technology Development Program of the NRF (NRF-2019M3E5D3073102 and NRF-2019R1H1A2039682), and a National IT Industry Promotion Agency (NIPA) grant (No. S0252-21-1001, Development of AI Precision Medical Solution (Doctor Answer 2.0).

\bibliography{main}
\bibliographystyle{icml2023}

\newpage
\appendix
\onecolumn
\section{Evaluation Model Selection}
\label{appendixA}

In order to select the best evaluation model for counterfactual estimation,  we conducted comparison experiments with three candidate model structures. For counterfactual estimation with neural networks, we aimed at a backbone model to output a balanced representation that is non-predictive to a factual prescription, reducing selection bias. Two prediction heads following the backbone model estimate prescription and A1c, respectively, from this balanced representation. As the EHR dataset covered in this paper has a high missing rate, we trained the three models to estimate an A1c value of the very next admission rather than autoregressive further estimation to avoid unreliable bias. The candidate models are as follows:

•  \quad Encoder part of Counterfactual Recurrent Network (CRN-E)

•  \quad Transformer with gradient reversal layer (T-GRL)

•  \quad Transformer with domain confusion loss (T-DC)

\vspace{0.5cm}

\textbf{CRN-E : }
Counterfactual Recurrent Network \cite{bica2020estimating} is an LSTM-based model that learns balanced representation via domain adaptation with Gradient Reversal Layer (GRL) \cite{ganin2016domain}. It consists of encoder and decoder parts; the encoder that infers the very next outcome from a balanced representation of previous history and the decoder that autoregressively predicts future outcomes. As the first candidate, CRN-E is the encoder part of the Counterfactual Recurrent Network.

Let $\Phi \left ( \bar{h}_{t} \right )$ be the balanced representation given at timestep ${t}$,  ${G}_{Y}$ the prediction head for A1c outcomes according to the balanced representation $\Phi \left ( \bar{h}_{t} \right )$ and a given prescription ${a}_{t}$, and ${G}_{A}$ be the prediction head for prescriptions as a domain classifier. $\theta_{Y}$ and $\theta_{A}$ denote the trainable parameters in ${G}_{Y}$ and ${G}_{A}$, and $\theta_{R}$ denotes the trainable parameters in the LSTM backbone. For time step ${t}$, let $\mathcal{L}_{t,y} \left (\theta_{Y},\theta_{R} \right )$ be a loss for the A1c outcome ${y}_{(t+1)}$ and $\mathcal{L}_{t,a} \left (\theta_{A},\theta_{R} \right)$ be a loss for medication set ${a}_{t}$, which are defined as follows:

\begin{center}
    $\mathcal{L}_{t,y} \left (\theta_{Y},\theta_{R} \right ) = {|| y_{t+1}- G_y \left ( \Phi \left ( \bar{h}_{t};\theta_R \right ), a_t ; \theta_Y \right )||}^2$ and
    \vspace{0.5cm}
    
    $\mathcal{L}_{t,a} \left (\theta_{A},\theta_{R} \right) = -\displaystyle\sum_{j=1}^{d_a}\mathbb{I}_{\{a_t=a_j\}}log\left(G_A\left ( \Phi \left ( \bar{h}_{t};\theta_R \right );\theta_A\right)\right)$,
\end{center}

where $\mathbb{I}_{\{.\}}$  is the indicator function and $d_a$ is the number of medication sets. With these losses, for the non-predictive  representation with respect to prescriptions, we trained all parameters of the CRN-E in the direction of minimizing $\mathcal{L}_{t,y}$ and maximizing $\mathcal{L}_{t,a}$ by using the GRL architecture. The overall loss is as follows:

\begin{center}
    $\mathcal{L}_{t} \left (\theta_{Y}, \theta_{A}, \theta_{R}\right) = \mathcal{L}_{t,y}\left (\theta_{Y},\theta_{R} \right ) - \alpha \mathcal{L}_{t,a} \left (\theta_{A},\theta_{R} \right)$
\end{center}

where the hyperparameter $\alpha$ represents the trade-off between the two predictions. Embedding layers for input features with missing values had the same structures as in the CDT.

\vspace{0.5cm}

\textbf{T-GRL : } The second model is Transformer with GRL, which uses the same representation balancing approach as the CRN-E except that a Transformer network replaces the LSTM. A backbone model of the T-GRL is GPT, and specific backbone architectures are similar to that of the CDT except for goal-related parts.

\vspace{0.5cm}

\textbf{T-DC : } T-DC also applies a domain adaptation approach to GPT similar to T-GRL.  Instead of GRL architecture, it uses domain confusion (DC) loss \cite{tzeng2015simultaneous} for balanced representation. The notations for losses are the same as in CRN-E, but let $\theta_R$ be the trainable parameters in the GPT backbone in T-DC. Assuming that the parameters of the $G_A$ are optimized to $\hat{\theta}_A$, $\theta_R$ is updated to prescription invariant direction with the DC loss, a cross-entropy of the uniform distribution over a categorial output of the $G_A$:

\begin{center}

    $\mathcal{L}_{t, conf} \left (\theta_{R}\right) = -\displaystyle\sum_{j=1}^{d_a}\frac{1}{d_a}\log\left(G_A\left(\Phi\left(\bar{h}_t; \theta_R\right);\hat{\theta}_A\right)\right)$

\end{center}

On the other hand, the $G_A$ is trained, assuming that the parameters of the GPT backbone are optimized to $\hat{\theta}_R$. In practical implementation, overall training of the T-DC is performed by iteratively minimizing the following losses: 

\begin{center}

    $\mathcal{L}_{t} \left (\theta_{Y},\theta_{R} \right ) =\mathcal{L}_{t,y}\left (\theta_{Y},\theta_{R} \right ) + \alpha\mathcal{L}_{t, conf} \left (\theta_{R}\right) $  and

    $\mathcal{L}_{t,a} \left (\theta_{A}\right) = -\displaystyle\sum_{j=1}^{d_a}\mathbb{I}_{\{a_t=a_j\}}log\left(G_A\left ( \Phi \left ( \bar{h}_{t};\hat{\theta}_R \right );\theta_A\right)\right)$,
\end{center}

where $\alpha$ is a balancing hyperparameter. \cref{table4} compares the structure of the three candidate models. 
\begin{table}[hbt!]
    \centering
    \caption {Feature difference between three candidate models.}
    \label{table4}
    \vspace{0.3cm}
    \renewcommand{\arraystretch}{1.2}
    \begin{tabular}{c c c c}
        \Xhline{3\arrayrulewidth}
         \textbf{Feature} & \textbf{CRN-E} & \textbf{T-GRL} & \textbf{T-DC} \\ \hline\hline
         Backbone Structure & LSTM  & Transformer(GPT) & Transformer(GPT) \\
         Representation Balancing Method & GRL & GRL & DC \\
         \Xhline{3\arrayrulewidth}
    \end{tabular}
\end{table}

We conducted hyperparameter tuning with training and validation datasets for all candidate models. The hyperparameters optimized for each model are shown in \cref{table5}. The hyphen symbol(-) means that the model does not use the corresponding hyperparameter. The lambda rate is a hyperparameter for reversed gradients during backpropagation.

\begin{table}[hbt!]
    \centering
    \caption {Hyperparameter of each evaluation model.} 
    \label{table5}
    \vspace{0.3cm}
    \setlength{\tabcolsep}{18pt}
    \renewcommand{\arraystretch}{1.2}
    \begin{tabular}{c | c c c}
        \Xhline{3\arrayrulewidth}
         \textbf{Hyperparameters} & \textbf{CRN-E} & \textbf{T-GRL} & \textbf{T-DC} \\ \hline\hline
         Iteration of Hyperparameter Search & 50  & 50 & 50 \\
         Early Stop Patience & 30 & 50 & 50 \\
         Minibatch Size & 64 & 128 & 128 \\
         Number of Layers & 4 & 3 & 4 \\
         Number of Medication Layers & 2	 & 1 & 3 \\
         Number of Attention Heads & - & 1 & 4 \\
         Activation Function of GPT & - & ReLU & ReLU \\
         Learning Rate  & 0.0001	 & 0.0001 & 0.0001 \\
         Warmup Steps & 500 & 500 & 500 \\
         Lambda Rate & 10 & 10 & - \\
         Balancing Alpha & 0.2 & 0.05 & 0.1 \\
         Dimension for Column Embedding & - & 64 & 32 \\
         Dimension for Value Embedding & 128 & 64 & 96 \\
         Gradient Normalization Clip & 1.0 & 1.0 & 0.5 \\
         Weight Decay & 0.01	& 0.0001 & 0.001\\
         \Xhline{3\arrayrulewidth}
    \end{tabular}
\end{table}

As we do not have access to the ground truth values for counterfactual estimation, we compared the estimation performance based on mean square errors on factual test data. The result of a comparison experiment with the tunned models is shown in \cref{table6}, and we selected CRN-E as an evaluation model for our proposed CDT.

\begin{table}[hbt!]
    \centering
    \caption {Comparison of the three candidate models.}
    \label{table6}
    \vspace{0.3cm}
    \setlength{\tabcolsep}{24pt}
    \renewcommand{\arraystretch}{1.3}
    \begin{tabular}{c c c c}
        \Xhline{3\arrayrulewidth}
         \textbf{Model} & \textbf{CRN-E} & \textbf{T-GRL} & \textbf{T-DC} \\ \hline\hline
         Mean Squared Error & \textbf{0.3767}  & 0.5907 & 0.6442 \\
         \Xhline{3\arrayrulewidth}
    \end{tabular}
\end{table}

\section{Hyperparameter Tuning for Clinical Decision Transformer}
\label{appendixB}

We have established a group of hyperparameter candidates referring to the hyperparameters used by the Decision Transformer \cite{chen2021decision}. Our hyperparameters for the CDT model are shown below in \cref{table7}. The Youden index\cite{youden1950index} on the validation set was used as a criterion for selecting the best hyperparameter set. In each iteration, the model was trained using the AdamW \cite{loshchilov2017decoupled} optimizer with early stopping conditioned on the cross-entropy loss.

\begin{center}
    \begin{table}[hbt!]
        \centering
        \caption{Hyperparameters of the CDT model.}
        \label{table7}
        \vspace{0.3cm}
        \setlength{\tabcolsep}{32pt}
        \renewcommand{\arraystretch}{1.2}
        \begin{tabular}{c c}
            \Xhline{3\arrayrulewidth}
             \textbf{Hyperparameters} & \textbf{Value} \\ \hline\hline
             Iteration of Hyperparameter Search & 50 \\
             Early Stop Patience	& 100 \\
             Minibatch Size	& 128 \\
             Number of Layers & 4 \\
             Number of Attention Heads	& 2\\
             Activation Function of GPT	& ReLU\\
             Learning Rate	& 0.0001\\
             Warmup Steps	& 1000\\
             Dimension for Column Embedding	& 16\\
             Dimension for Value Embedding	& 112\\
             Gradient Normalization Clip &	0.25\\
             Weight Decay	& 0.0001\\
             \Xhline{3\arrayrulewidth}
        \end{tabular}
    \end{table}
\end{center}

\section{Behavior Cloning Model}
\label{appendixC}

We built a behavior cloning model upon LSTM networks. As in the CDT training, the behavior cloning model was also trained by using a cross-entropy loss over factual medication sets prescribed by clinicians. Unlike the CDT, this model aims to imitate the clinician’s behavior and is not allowed to manipulate recommendations with any intention. Hyperparameters are shown in \cref{table8}.

\begin{center}
    \begin{table}[hbt!]
        \centering
        \caption{Hyperparameters of the behavior cloning model.}
        \label{table8}
        \vspace{0.3cm}
        \setlength{\tabcolsep}{32pt}
        \renewcommand{\arraystretch}{1.3}
        \begin{tabular}{c c}
            \Xhline{3\arrayrulewidth}
             \textbf{Hyperparameters} & \textbf{Value} \\ \hline\hline
             Iteration of Hyperparameter Search & 50 \\
             Early Stop Patience	& 30 \\
             Minibatch Size	& 64 \\
             Number of Layers & 3 \\
             Learning Rate	& 0.0001\\
             Dimension for Embedding	& 128\\
             Gradient Normalization Clip &	0.25\\
             \Xhline{3\arrayrulewidth}
        \end{tabular}
    \end{table}
\end{center}

\section{Diabetes Dataset}
\label{appendixD}

This data extraction from an EHR system was approved by an official review committee of the National Health Insurance Corporation of Korea and the Institutional Review Board of the Korea University Ansan Hospital (2021AS0264). We screened 4788 patients diagnosed with diabetes who were admitted between January 2015 and July 2021. Detailed information is summarized in \cref{table9,table10,table11,table12,table13,table14}.

\begin{table}[hbt!]
    \centering
    \caption{Baseline characteristics of the extracted dataset.}
    \label{table9}
    \vspace{0.3cm}
    \setlength{\tabcolsep}{32pt}
    \renewcommand{\arraystretch}{1.3}
    \begin{tabular}{c c}
        \Xhline{3\arrayrulewidth}
         \textbf{Features} &  \textbf{Value} \\ \hline\hline
         Number of Patients	& 4788 \\
        Shortest Admission Length &	1 \\
        Longest Admission Length &	41 \\ 
        Mean Admission Length	& 13.64 \\
        Median Admission Length	& 13.0 \\
        Number of Baseline Features	& 14 \\
        Number of Laboratory Features	& 18 \\
        Number of Prescribable Medications	& 12 \\
         \Xhline{3\arrayrulewidth}
    \end{tabular}
\end{table}

\begin{table}[hbt!]
    \centering
    \caption{Characteristics of the dataset for CDT after the goal-conditioned sequencing process.}
    \label{table10}
    \vspace{0.3cm}
    \setlength{\tabcolsep}{32pt}
    \renewcommand{\arraystretch}{1.3}
    \begin{tabular}{c c}
        \Xhline{3\arrayrulewidth}
         \textbf{Features} &  \textbf{Value} \\ \hline\hline
        Number of Subsequences	& 9046 \\
        Shortest Admission Length &	1 \\
        Longest Admission Length	& 34 \\
        Mean Admission Length &	5.12 \\
        Median Admission Length	& 3.0 \\
         \Xhline{3\arrayrulewidth}
    \end{tabular}
\end{table}

\begin{table}[hbt!]
    \centering
    \caption{Characteristics of the dataset for the evaluation model after removing non-containing A1c data.}
    \label{table11}
    \vspace{0.3cm}
    \setlength{\tabcolsep}{32pt}
    \renewcommand{\arraystretch}{1.3}
    \begin{tabular}{c c}
        \Xhline{3\arrayrulewidth}
         \textbf{Features} &  \textbf{Value} \\ \hline\hline
        Number of Sequences	& 4091 \\
        Shortest Admission Length	& 1\\
        Longest Admission Length	& 38\\
        Mean Admission Length	& 11.78\\
        Median Admission Length	& 12.0\\
         \Xhline{3\arrayrulewidth}
    \end{tabular}
\end{table}

\begin{table}[hbt!]
    \centering
    \caption{Prescribable medications.}
    \label{table12}
    \vspace{0.3cm}
    \setlength{\tabcolsep}{28pt}
    \renewcommand{\arraystretch}{1.3}
    \begin{tabular}{c c}
        \Xhline{3\arrayrulewidth}
         \textbf{Features} &  \textbf{Type of Feature} \\ \hline\hline
        Metformin	& Binary\\
        Sulfonylurea	& Binary\\
        Thiazolidinedione	& Binary\\
        DPP-4 Inhibitor	& Binary\\
        Insulin	& Binary\\
         SGLT2 inhibitor	& Binary\\
        Glinide	& Binary\\
        $\alpha$-Glucosidase Inhibitor	& Binary\\
        GLP-1 Receptor Agonist	& Binary\\
        Hypertension Medication	& Binary\\
        Hyperlipidemia Medication	& Binary\\
        Diabetes Mellitus Neuropathy Medication	& Binary\\
         \Xhline{3\arrayrulewidth}
    \end{tabular}
\end{table}

\begin{table}[hbt!]
    \centering
    \caption{Characteristics of the baseline columns.}
    \label{table13}
    \vspace{0.3cm}
    \setlength{\tabcolsep}{38pt}
    \renewcommand{\arraystretch}{1.3}
    \begin{tabular}{c c}
        \Xhline{3\arrayrulewidth}
         \textbf{Features} &  \textbf{Type of Feature} \\ \hline\hline
        Sex	& Binary\\
        Age	& Categorical(4--120)\\
        Ischemic Heart Disease	& Binary\\
        Congestive Heart Failure	&  Binary\\
        Arrhythmia	&  Binary\\
        Ischemic Stroke	&Binary\\
        Hemorrhagic Stroke	& Binary\\
        Diabetes Mellitus Retinopathy	& Binary\\
        Kidney Transplant	& Binary\\
        Hemodialysis	& Binary\\
        Peritoneal Dialysis	& Binary\\
        Hypertension	& Binary\\
        Dyslipidemia	& Binary\\
        Diabetes Mellitus Neuropathy	& Binary\\
        \Xhline{3\arrayrulewidth}
    \end{tabular}
\end{table}

\begin{table}[hbt!]
    \centering
    \caption{Characteristics of the laboratory test columns.}
    \label{table14}
    \vspace{0.3cm}
    \setlength{\tabcolsep}{32pt}
    \renewcommand{\arraystretch}{1.3}
    \begin{tabular}{c c}
        \Xhline{3\arrayrulewidth}
         \textbf{Features} &  \textbf{Missing Rate (\%)} \\ \hline\hline
        Glucose	& 2.70\\
        Hemoglobin A1c	& 37.30\\
        Cholesterol	& 7.22\\
        High-Density Lipoprotein Cholesterol	& 7.35\\
        Low-Density Lipoprotein Cholesterol	& 99.59\\
        Triglyceride	 & 7.15\\
        Aspartate aminotransferase	& 20.18\\
        Alanine aminotransferase	& 20.18\\
        Albumin	& 22.83\\
        Microalbumin	& 70.71\\
        Microalbumin (Urine, CSF)	& 100.00\\
        Microalbumin/Creatinine in Urine	& 70.71\\
        Microalbumin (RIA)	& 99.26\\
        Blood Urea Nitrogen	& 21.92\\
        Creatinine (Serum)	& 21.63\\
        Creatinine (Urine)	& 70.71\\
        Creatinine (BF, Urine)	& 98.69\\
        Bilirubin	& 20.94\\
         \Xhline{3\arrayrulewidth}
    \end{tabular}
\end{table}

\end{document}